# Distributed Anytime MAP Inference

Joop van de Ven
Australian Centre for Field Robotics
School of Information Technologies
The University of Sydney
Sydney 2006, NSW, Australia
j.vandeven@acfr.usyd.edu.au

Fabio Ramos
Australian Centre for Field Robotics
School of Information Technologies
The University of Sydney
Sydney 2006, NSW, Australia
f.ramos@acfr.usyd.edu.au

## Abstract

We present a distributed anytime algorithm for performing MAP inference in graphical models. The problem is formulated as a linear programming relaxation over the edges of a graph. The resulting program has a constraint structure that allows application of the Dantzig-Wolfe decomposition principle. Subprograms are defined over individual edges and can be computed in a distributed manner. This accommodates solutions to graphs whose state space does not fit in memory. The decomposition master program is guaranteed to compute the optimal solution in a finite number of iterations, while the solution converges monotonically with each iteration. Formulating the MAP inference problem as a linear program allows additional (global) constraints to be defined; something not possible with message passing algorithms. Experimental results show that our algorithm's solution quality outperforms most current algorithms and it scales well to large problems.

## 1 Introduction

Undirected graphical models are powerful tools for modelling many real world problems. They have been successfully applied to a diverse set of problems such as: image processing [1], protein design [2], and text labelling [3]. One desirable operation on such models is to infer their most probable configuration; the maximum a posteriori (MAP) problem. For tree-structured graphs algorithms exist that are guaranteed to compute the globally optimal MAP solution in polynomial time (see for example: [4, 5]).

Finding the MAP solution for arbitrary graphs has been proven to be NP-hard [6]. For such graphs approximation algorithms are required to generate solutions in a feasible time-span. One particularly popular algorithm is based on Max-Product Belief Propagation due to Pearl [5]. Max-Product is exact for tree-structured graphs. For arbitrary graphs the algorithm has been adapted such that it runs for a number of iterations and is known as Loopy Belief Propagation (LBP, see [7]). LBP has (at best) weak guarantees on convergence and optimality, i.e. local optimality or guarantees for specific types of loopy graphs. Despite this, the algorithm has been shown to generate good results for a large number of problems.

In this paper we propose a novel algorithm for the MAP inference problem of graphical model $G = (V, E)$. Starting from a quadratic formulation over the nodes, $s \in V$ (analogous to [8]), the problem is transformed into an integer formulation over the edges, $(s, t) \in E$. It is subsequently relaxed into a Linear Program (LP). The transformation from quadratic to linear increases both the number of variables ($O(|V|k) \to O(|E|k^2)$) as well as constraints ($O(|V|) \to O(|E| + \sum_{s \in V}(|\mathcal{N}(s)| - 1)k)$). Where for ease of notation each node has $k$ states, while $\mathcal{N}(s)$ are the neighbours of node $s$. The resulting LP formulation is equivalent to the standard MAP LP formulation (see for example [9]). However, defining the LP over the edge variables has certain advantages as discussed next.

For problems where the proposed LP relaxation fits into memory any LP solver may be used to compute the optimal solution directly. For medium to large-scale problems there may not be sufficient memory available to solve the LP. Our algorithm is particularly suited for these cases as we explore the structure of the constraints to allow the application of the Dantzig-Wolfe decomposition principle [10]. The decomposition principle reformulates the LP into a number of subprograms, one for each edge in our case, together with a master program. The subprograms can be solved independently and distributed, while the master program solves for the optimal solution using a significantly reduced number of variables. The algorithm is solved iteratively, solutions to the subprograms are used to update the master program and vice versa. At each iteration the decomposition principle guarantees that the solution will be closer to the optimal solution, i.e. an *anytime* algorithm.

The advantages of our algorithm are: 1) it is able to solve very complex problems that few other algorithms are able to solve; 2) it scales well for large problems and allows the use of linear programming even for graphs where the state space does not fit in memory; 3) it can be distributed and effectively use the multicore hardware currently available and; 4) it allows the definition of global constraints which are difficult to enforce in message-passing based algorithms. This is important for a number of real-world problems involving graph matching and data association.

## 2  Related Work

Several variants of Max-Product Belief Propagation have been proposed. Generalised Belief Propagation [11] extends the message passing from pairs of connected nodes to higher order cliques resulting in better approximations. Tree-Reweighted Max-Product methods (TRW, [12, 13]) on the other hand decompose the original graph into a convex combination of tree-structured graphs. The tree-structured graphs guarantee efficient computations, while the convex combination allows the computation of an upper bound on the optimal solution. However, Yanover et al. [14] showed that TRW fails to solve the problems used in the experiments of section 5.1.

TRW has strong connections to the Max-Product Linear Programming (MPLP) algorithm proposed by Globerson and Jaakkola [15]. MPLP is defined as a block coordinate descent in the dual of a LP relaxation constrained by the local marginal polytope. MPLP has all the advantages of message passing algorithms but it also has strong convergence guarantees. However, as the problem is solved in the dual, it means optimising an upper bound and not the MAP problem directly. As a result, MPLP is also unable to solve most of the problems used in the experiments (see section 5.1). The method developed by Sontag et al. [2, 16] is considered the state of the art in MAP inference. It extends MPLP by iteratively adding clusters to the MPLP formulation thus approximating the marginal polytope.

Komodakis et al. [17] solve the MAP problem by decomposition. Starting from the same LP relaxation as MPLP, the problem is transformed into its unconstrained dual Lagrangian using Lagrange multipliers. In the dual the problem is decomposed, where each sub problem is formed by a spanning tree (similar to TRW methods). The solutions of the sub problems are used in a projected sub-gradient method to update the potential values. However, their method (Dual Decomposition, DD) requires all potential values to be updated and communicated to each sub problem. Thus making this approach sub-optimal for problems with large state spaces.

Ravikumar and Lafferty [8] formulate the MAP problem as a Quadratic Program (QP) relaxation. QP relaxations are a more natural fit to the MAP problem as pair-wise potentials are quadratic in nature. For many practical problems the QP relaxation is non-convex, thus requiring further approximations in order to produce an algorithm that is solvable in polynomial time. The drawback to using a QP relaxation is that it requires memory squared in the number of states. So for medium to large scale problems this approach quickly becomes impractical. In addition, the QP relaxation has been shown to generate poorer results compared to a LP relaxation [18].

More recently, Kumar and Zilberstein [19] approached the MAP estimation problem with an interesting mean field approximation method. Their method approximates the problem by considering only distributions that factorise as a product of distributions over individual nodes. The resulting non-convex problem is represented by an equivalent mixture of Bayes nets with one network for each edge. Expectation-Maximisation (EM, [20]) is subsequently used to derive a message passing algorithm. The EM message passing algorithm is computationally efficient but sensitive to initial conditions.

The performance of message passing algorithms degrades significantly for large scale problems. Due to memory restrictions it may not be possible to keep all pair-wise potentials in memory. In such cases, potentials will have to be recomputed with each message sent. This is not an algorithmic issue but a practical one, nonetheless it adds to the computational cost. In addition large messages need to be constructed, again increasing the cost. Our approach does not suffer this limitation as it is intrinsically distributed. Edge potentials are computed only once for each distributed subprogram. Furthermore, our approach permits specification of constraints on the solution; something not possible with the above methods. Finally, we show that our algorithm is able to solve graphs that TRW methods and MPLP ([12, 13] and [15] respectively) are unable to solve, since our method solves the primal directly rather than optimise a bound.

## 3 LP Formulation

An undirected graphical model $G = (V, E)$ represents a probability distribution $p_G(x_1, \ldots, x_N)$ over $N = |V|$ variables. The vertices $s \in V$ of the graph index the random variables $x_s$ of the distribution, while the edges $(s, t) \in E$ of the graph capture relationships between variables $x_s$ and $x_t$. Let $\mathcal{C}$ be the set of all cliques of the graph. The distribution must factor as a product of clique potentials $\phi_c(X_c)$, where $c \in \mathcal{C}$ and $X_c$ are the clique's variables. Yedidia et al. [11] showed that, without loss of generality, it is possible to assume that the graph is a pair-wise Markov Random Field, i.e. the set of cliques $\mathcal{C}_M = \{(s,t) \in E\}$. As a result, the log of the distribution of $X = \{x_1, \ldots, x_N\}$, for the graph $G$ with potentials $\Phi = \{\phi_c | c \in \mathcal{C}_M\}$ is given by:

$$\log p_G(X; \Phi) = \sum_{s \in V} \phi_s(x_s) + \sum_{(s,t) \in E} \phi_{st}(x_s, x_t) - C, \qquad (1)$$

where $C$ is the log of the partition function. In the remainder we only consider variables $x_s$ that take on values from a finite discrete set $\mathcal{X}_s$. Each $x_s$ is a vector of length $|\mathcal{X}_s|$ with elements $x_s^i \in \{0, 1\}$ and $\sum_i x_s^i = 1$. Furthermore, without loss of generality we include the local potentials in the pair-wise potentials. For discrete graphical models the combined pair-wise potential may then be expressed as:

$$Q_{st} = \phi_{st} + (e\phi_s^T)/|\mathcal{N}(s)| + (\phi_t e^T)/|\mathcal{N}(t)|, \qquad (2)$$

where $\mathcal{N}(s)$ is the set of neighbours of node $s$. Division by the number of neighbours distributes the local potential evenly over the pair-wise potentials while leaving the MAP value unaltered. The vector $e$ is an appropriately sized vector of 1s. This leads to the following quadratic integer MAP problem,

$$\begin{aligned} X^{MAP} &= \operatorname*{argmax}_X \log p_G(X; \Phi) \\ &= \operatorname*{argmax}_X \sum_{(s,t) \in E} x_s^T Q_{st} x_t. \end{aligned} \qquad (3)$$

For small scale problems equation 3 may be solved by a QP relaxation. However, for medium or large scale problems the resulting relaxation will quickly become too large to fit into memory. Instead we reformulate the quadratic objective function into a LP by substitution. The substitution transforms the problem from an optimisation over the nodes into one over the edges with a constraint structure that allows the LP to be solved in a distributed manner.

For each edge $(s, t) \in E$ define the edge variable

$$\begin{aligned} y_{st} = (x_s^i x_t^j | i = 1, \ldots, |\mathcal{X}_s|, \\ j = 1, \ldots, |\mathcal{X}_t|, (s, t) \in E) \end{aligned} \qquad (4)$$

which, by the discrete nature of $x_s$ and $x_t$, has elements $y_{st}^k \in \{0, 1\}$ and $\sum_k y_{st}^k = 1$ ($k = 1, \ldots, |\mathcal{X}_s||\mathcal{X}_t|$). Equally, the cost $c_{st}$ can be constructed from $Q_{st}$ by ordering the elements of $Q_{st}$ into a vector corresponding to the elements of $y_{st}$. The LP relaxation to the MAP problem is then formulated as:

$$\begin{aligned} &\text{Maximise} \\ &\quad \sum_{(s,t) \in E} c_{st}^T y_{st} \\ &\text{Subject to} \\ &A_{st} y_{st} - A_{su} y_{su} = 0 \quad \forall s \in V, t \in \mathcal{N}(s), \\ &\qquad\qquad\qquad\qquad\qquad \forall u \in \mathcal{N}(s) \setminus t \\ &\sum_k y_{st}^k = 1 \quad \forall (s, t) \in E \\ &0 \leq y_{st}^k \leq 1 \quad \forall (s, t) \in E, \\ &\qquad\qquad\qquad k = 1, \ldots, |\mathcal{X}_s||\mathcal{X}_t|. \end{aligned} \qquad (5)$$

The constraints defined by the matrix coefficients $A_{st}$ are the consistency constraints. The elements of $A_{st}$ are 0 or 1 such that $A_{st}^{i,\bullet} y_{st} = x_s^i$, where $A_{st}^{i,\bullet}$ is the $i$-th row of $A_{st}$. Consistency constraints are discussed in more detail in section 3.1. The second set of constraints express a uniqueness of solution for each edge; only one element of the edge variable $y_{st}$ may be active. While the third set of constraints capture the relaxation from an integer program to a linear program.

Equation 5 is equivalent to standard MAP LP formulation (see for example [9]). The advantage of using equation 5 over the standard formulation is that it has fewer variables and constraints. The reduction in variables is straightforward since local potentials (variables) are included in the pair-wise variables. The constraints are less obvious but more important. As shall be shown in section 4, the communication overhead is proportional to the number of constraints; fewer constraints are preferable. For problems with large state space and small treewidth, equation 5 has significantly fewer constraints compared to variables (section 5.2 contains an example). As a result the proposed method will have a smaller communications overhead compared to, for example, the Dual Decomposition method [17].

### 3.1 Solution Consistency

Equation 5 is defined over the edges. As such constraints are required to ensure the solution remains consistent in the node variables.

**Definition** Let $m_{s|t} = A_{st} y_{st}$ be a *marginal*, for node variable $x_s$, of the edge variable $y_{st}$.

**Proposition 3.1** *The solution for the edge variables $\{y_{st} | \forall t \in \mathcal{N}(s)\}$ is consistent in the node variable $x_s$ when the marginals $\{m_{s|t} | \forall t \in \mathcal{N}(s)\}$ are all equal.*

**Proof** The proof can be obtained by simple substitution of $y_{st}^k = x_s^i x_t^j$ and $\sum_i x_s^i = 1$. ∎

Consistency constraints are specified over pairs of edges, i.e. as the difference between pairs of marginals $m_{s|t}$ and $m_{s|u}$. For a given node $s$ one edge is used as the reference edge; edge $(s,t)$ in equation 5. All consistency constraints are specified relative to the reference edge resulting in a minimum of constraints generated. Subsequently solving equation 5 will result in a solution for the edge variables. The mapping from $y_{st}$ to the node variables $x_s$ is given by the following proposition.

**Proposition 3.2** *If the linear program of equation 5 has a feasible solution, then the mapping from $y_{st}$ to $x_s$ is given by $x_s = m_{s|t}$ for any $t \in \mathcal{N}(s)$.*

**Proof** The equality $x_s = m_{s|t}$ follows directly from the definitions of $A_{st}$, $y_{st}$ and $m_{s|t}$. Proposition 3.1 permits any $t \in \mathcal{N}(s)$ provided the solution is consistent. Solution consistency, and therefore proposition 3.1, is ensured by virtue of a feasible solution; all constraints are met. ∎

### 3.2 Additional Requirements

Certain optimisation problems have additional requirements (or constraints) imposed on them. To ensure neighbouring nodes $x_s$ and $x_t$ have distinct (or equality) solutions for states $i$ and $j$ respectively, add one of the following constraints to equation 5 for *each* such state:

$$\begin{array}{ll} D_{st,ij} y_{st} \leq 1 & \forall (s,t) \in E, \forall (i,j): x_s^i \neq x_t^j \\ E_{st,ij} y_{st} = 0 & \forall (s,t) \in E, \forall (i,j): x_s^i = x_t^j, \end{array} \quad (6)$$

where $D_{st,ij} = A_{st}^{i,\bullet} + A_{ts}^{j,\bullet}$ ensures a distinct solution, while $E_{st,ij} = A_{st}^{i,\bullet} - A_{ts}^{j,\bullet}$ ensures an equal solution between the states $i$ and $j$ of nodes $x_s$ and $x_t$. Note that the constraints of equation 6 are defined locally. This approach can easily be extended to global constraints. However, this is omitted for brevity.

## 4 Decomposition

The Dantzig-Wolfe decomposition principle [10] allows a LP with a special block-matrix structure to be broken up into a number of independent subprograms. The subprograms are iteratively adjusted to take into account global state (simplex multipliers) due to a master program. The reader is referred to [10, Chapter 10] for a detailed discussion and proofs. In this section we provide an interpretation of the principle in the context of MAP inference for graphical models.

The block-angular system,

$$\begin{array}{l} \text{Maximise} \\ \quad c_1^T y_1 + \ldots + c_K^T y_K \\ \text{Subject to} \\ \quad B_1 y_1 + \ldots + B_K y_K = b \\ \quad F_1 y_1 \qquad\qquad\qquad\quad = f_1 \\ \quad \ddots \qquad\qquad\qquad\qquad \vdots \\ \qquad\qquad\qquad F_K y_K = f_K \\ \quad y_i \geq 0 \qquad\qquad\quad i = 1, \ldots, K, \end{array} \quad (7)$$

allows decomposition to be applied to the MAP problem for $K = |E|$. The matrices $\{B_i | i = 1, \ldots, K\}$ form the coupling constraints, they capture interactions between subprograms. The constraints unique to each subprogram are constructed from $\{F_i | i = 1, \ldots, K\}$.

The constraints of equation 5 have this block-angular structure. $B_{st}$ is constructed from the set $\{A_{st}\}$ for edge $(s,t)$, appropriately padded with 0s to ensure all $B$ have the same number of rows. If additional requirements are specified (see section 3.2) then the sets $\{D_{st}\}$ and $\{E_{st}\}$ are also included in $B_{st}$. The subprogram constraints are the uniqueness of solution constraints; $F_{st} = \sum y_{st}$.

The decomposition principle exploits the Resolution Theorem [10]. Briefly, the Resolution Theorem states that every feasible solution of the convex polyhedral set $Ax = b, x \geq 0$ can be represented as a convex combination of its extreme points[1] (see [10, Theorem 10.5] for more details). Using the Resolution Theorem, equations of the form of equation 7 can be transformed into a master program together with $K$ subprograms. The master program maximises a convex combination of extreme points, while the subprograms generate extreme points at each iteration.

We now present the steps of the algorithm, each of the steps are discussed in more detail in the sections to follow:

1. Initialise the algorithm (section 4.1) to find an initial basic feasible solution.

2. Solve the master program using columns corresponding to the initial basic feasible solution. This provides global state in the form of the simplex multipliers $(\pi, \gamma)$; see section 4.3 for more details.

3. Solve all subprograms using the current simplex multipliers (section 4.2).

---
[1] Normalised extreme homogeneous solutions are omitted as our subprograms cannot generate these.

4. Add columns to the master program according to optimality of subprogram solutions and corresponding column cost. Solve the master program to obtain new simplex multipliers $(\pi, \gamma)$; section 4.3.

5. If the master program has found the optimal solution go to step 6, if not go to step 3.

6. Transform the master program's solution to the solution of equation 5 and perform rounding if required, see section 4.4 for more details.

### 4.1 Initialisation

The aim of initialisation is to find an initial basic feasible solution. One common approach to initialising Dantzig-Wolfe decomposition is using a Simplex Phase 1 approach [10, Section 10.2.4]. This involves finding the maxima of each subprogram using the actual costs. The resulting solutions are used to start the master program. However, in our case the subprograms are trivial. This generally means that the consistency constraints are violated, thus preventing the decomposition from even starting. Instead when no additional requirements are specified, the procedure of algorithm 1 can be used to find an initial basic feasible solution in *one step*.

---
**Algorithm 1** Pseudo-code of algorithm initialisation.

1: **Input:** Graph $G = (V, E)$, potentials $\phi_s, \forall s \in V$ and $\phi_{st}, \forall (s,t) \in E$
2: **Output:** Initial basic feasible solution $\{\tilde{y}_{st} | \forall (s,t) \in E\}$
3: **for** $s \in V$ **do**
4: $\quad \tilde{\phi} \leftarrow \phi_s$
5: $\quad$ **for** $t \in \mathcal{N}(s)$ **do**
6: $\quad\quad \tilde{\phi} \leftarrow \tilde{\phi} + \sum_{x_t} \phi_{st}$
7: $\quad$ **end for**
8: $\quad \tilde{x}_s \leftarrow \text{argmax}_{x_s}(\tilde{\phi})$
9: **end for**
10: **for** $(s,t) \in E$ **do**
11: $\quad \tilde{y}_{st} \leftarrow (\tilde{x}_s^i \tilde{x}_t^j | i = 1, \ldots, |\mathcal{X}_s|, j = 1, \ldots, |\mathcal{X}_t|)$
12: **end for**

---

As can be seen from algorithm 1, a node's initial solution $\tilde{x}_s$ is found by maximising over the sum of local and marginalised pair-wise potentials. Once the initial solutions for each node have been found, they are mapped to their equivalent subprogram initial basic feasible solutions $\tilde{y}_{st}$. Since the initial solutions $\tilde{y}_{st}$ are based on node solutions, the consistency constraints are always met. The subprograms' initial basic feasible solutions are subsequently used to get the decomposition master program started.

When additional requirements (see section 3.2) are specified the above procedure is not guaranteed to find an initial basic feasible solution. In such cases one can adjust line 8 such that it takes account of additional requirements. If this is not possible then an initial solution will have to be obtained via other means. This requires replacing the **for**-loop on line 3. For example, many solution constraints allow a trivial solution, $\tilde{x}_s$ can be initialised with this trivial solution. In the more general case, algorithms that solve Constraint Satisfaction Problems (see for example [21]) can be used to find an initial solution for $\tilde{x}_s$.

### 4.2 Subprogram

For inference in a graph, the subprograms maximise a linear program over the edges as follows:

$$\begin{aligned} &\text{Maximise} \\ &\quad c_{st}^T y_{st} - B_{st}^T \pi \\ &\text{Subject to} \\ &\quad \sum_k y_{st}^k = 1 \\ &\quad 0 \leq y_{st}^k \leq 1 \quad k = 1, \ldots, |\mathcal{X}_s||\mathcal{X}_t|. \end{aligned} \quad (8)$$

The objective function of equation 8 is the actual cost of the edge, $c_{st}$, adjusted by the current state of the interactions, $B_{st}^T \pi$. Here $B_{st}^T$ are the concatenated consistency constraints (and optional additional requirements) while $\pi$ are the corresponding simplex multipliers (see section 4.3). This adjusted cost finds the edge's maximum based on the current global state of the algorithm. It is however not necessary to invoke a LP solver for each subprogram. There are two constraints for each edge, these express the uniqueness of solution ($y_{st}^k \in \{0,1\}$ and $\sum_k y_{st}^k = 1$). A solution to the subprograms can therefore be found by a simple maximisation over a vector; i.e. the solution is always an extreme point.

Let $\hat{y}_{st,i}$ represents the optimal solution for edge $(s,t)$ at iteration $i$. If $c_{st}^T \hat{y}_{st,i} - B_{st}^T \pi \neq \gamma^{st}$ then this solution is globally sub-optimal and it may be incorporated into the master program, provided it has not previously been incorporated. In case of a tie (multiple solutions with the same maximum) standard simplex tie breaking rules can be applied. In the experiments we select either the solution with the lowest index $k$ (analogous to *Bland's rule* [22]), or the index $k$ for which the actual cost is maximal (analogous to the *Largest-Coefficient rule* [23]).

### 4.3 Master Program

The purpose of the master program is twofold. First, it generates global state in the form of the simplex multipliers $\pi$ and $\gamma$. Second, any feasible solution to the master program can be transformed into a solu-tion of the original LP, equation 5. At each iteration of the algorithm columns are added to the master program depending on optimality of subprogram solutions (see section 4.2). The added columns allow the master program to update the simplex multipliers based on subprogram solutions.

For the MAP inference problem, the master program is defined as shown in equation 9.

$$\begin{aligned}
\text{Maximise} & \quad \sum_{(s,t)\in E}\sum_{i\in L_{st}} g_{st}^i \alpha_{st}^i \\
\text{Subject to} & \\
& \sum_{(s,t)\in E}\sum_{i\in L_{st}} G_{st}^{\bullet,i} \alpha_{st}^i = 0 \\
& \sum_{i\in L_{st}} \alpha_{st}^i = 1 \quad \forall (s,t) \in E \\
& \alpha_{st}^i \geq 0 \quad \forall (s,t) \in E, \\
& \qquad\qquad \forall i \in L_{st},
\end{aligned} \quad (9)$$

where $L_{st}$ is the set of iteration indices, of edge $(s,t)$, for which columns have been added to the master program. The element $g_{st}^i = c_{st}^T \hat{y}_{st,i}$ is the cost of the subprogram solution at iteration $i$. While $G_{st}^{\bullet,i} = B_{st}\hat{y}_{st,i}$ is the column of corresponding consistency constraints. The elements $\alpha_{st}^i$ are the convexity variables from the Resolution Theorem. They have a particularly elegant interpretation for the MAP problem; they represent the likelihood of the subprogram solutions $\hat{y}_{st,i}$. Note that $\pi$ are the simplex multipliers corresponding to the consistency constraints while $\gamma$ are the simplex multipliers of the convexity variables.

Up to $|E|$ columns may be added to the master program, one for each sub-optimal subprogram, at each iteration. However, not all of these prospective columns will aid in finding a solution. Quite to the contrary, often they will add unnecessary complexity to the master program. Instead of adding all prospective columns to the master program, a limited number of columns may be added at each iteration. In such cases columns are selected based on their cost. At iteration $i$ only columns corresponding to maximal costs $g_{st}^i$ are added (similar to pivoting in the simplex method).

With each iteration the master program will grow in size. It is possible that, after a number of iterations, the master program will become too complex for the solver to find a solution efficiently. In such situations the master program can be reduced in size and complexity simply by removing non-basic columns. Removal of non-basic columns does not impact on the solution quality as a LP solves a convex optimisation problem. It will however impact on the number of iterations until convergence. With more columns available the decomposition is able to find the optimal solution in fewer iterations, but it may take longer to solve each individual iteration.

Column generation procedures, such as Dantzig-Wolfe Decomposition, have the desirable property that both the size as well as the growth of the master program can be controlled. As a result even very large problems can be efficiently dealt with.

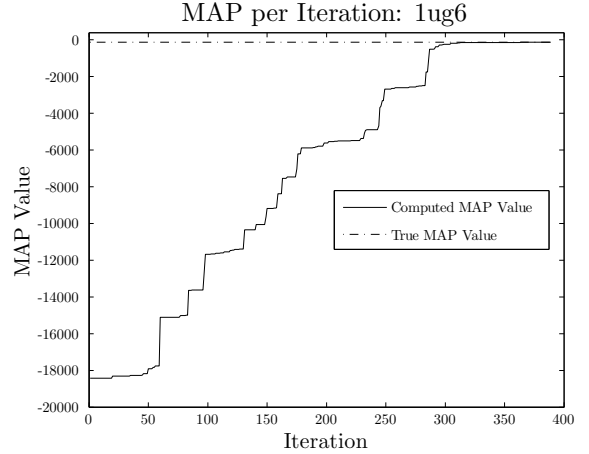

Figure 1: Monotonic convergence property of the master program for the Side-Chain Prediction graph 1ug6.

### 4.4 Optimal Solution

Dantzig-Wolfe decomposition is guaranteed to converge in a finite number of iterations [10, Theorem 10.4]. Once converged, the solution to equation 5 can be found from the convexity variables $\alpha_{st}^i$ and the corresponding subprograms' optimal solutions $\hat{y}_{st,i}$:

$$\check{y}_{st} = \sum_{i \in L_{st}} \alpha_{st}^i \hat{y}_{st,i}. \quad (10)$$

The globally optimal solution $\check{y}_{st}$ is the sum of the subprograms' optimal solutions scaled by their corresponding convexity variables (or likelihoods). The $\check{y}_{st}$ can be mapped back to a corresponding optimal node solution $\check{x}_s$ (see section 3.1).

Rounding may have to be applied to $\check{x}_s$ to find an integer solution. Instead of applying rounding schemes such as [24, 25], we construct an Integer Program (IP) over the non-zeros solution states of $\check{x}_s$. For each graph the percentage of nodes for which there is no integer solution is generally quite low (less than 5% in the experiments). These fractional nodes generate a set of probable answers. Analysis also showed that often one of these probable answers is the true optimal solution.

The IP operates on the same graph but on only those states $i$ (and corresponding potential values) for which $\check{x}_s^i \neq 0$. The resulting IP is small and can be solved in a fraction of the time of the decomposition (typically less than 10% of the running time). The benefit of using an IP is that it is guaranteed to *always* find the best solution out of the set of probable answers.

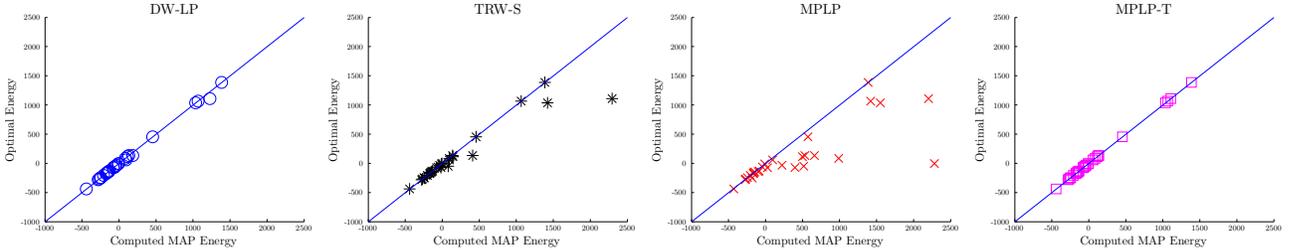

Figure 2: Side-Chain Prediction results. Each figure displays the computed MAP value versus its optimal value for all 30 graphs for a given algorithm. Points on the main diagonal are graphs for which an optimal solution is obtained while the horizontal deviation from the diagonal gives a measure of the MAP error. Left to right: proposed algorithm (DW-LP), TRW-S, MPLP, MPLP-T.

## 5 Experiments

### 5.1 Side-Chain Prediction

The performance of the proposed algorithm (DW-LP) is measured on the Rosetta Side-Chain Prediction data set [14, 2]. This involves finding the three-dimensional configuration of rotamers given the backbone structure of a protein [14]. Following Sontag et al. [2] we apply our algorithm to the 30 graphs that TRW [12] is unable to solve. We compare our algorithm against TRW-S (which improves on TRW, see [13]) and MPLP [15] as both, like our algorithm, consider only the local marginal polytope. In addition we also show the result for MPLP with tightening [2]. This is the current state of the art, we refer to it as MPLP-T.

Both TRW-S and MPLP are run for 1000 iterations or until convergence, whichever comes first. MPLP-T is operated as described in [2]. Like TRW-S and MPLP our algorithm is run for a maximum of 1000 iterations or until convergence. In addition up to 200 columns corresponding to maximal costs are added to the DW-LP master program at each iteration. While no non-basic columns are removed from the master program (see section 4.3).

Once finished the MAP value is computed from the node assignments. The MAP values are compared against their true values, differences smaller than 1e-6 are considered equal. The comparisons are given in table 1 and figure 2.

As can be seen from table 1 and figure 2, a like-for-like comparison shows that the proposed algorithm significantly outperforms both TRW-S as well as MPLP. It is able to solve more than half of the graphs to their optimal solution while the MAP error is much smaller in comparison. Note also that for all 30 graphs our algorithm converged to a solution within 1000 iterations (TRW-S and MPLP did not converge for all graphs). MPLP-T is able to solve all the problems as it iteratively adds triplet clusters to the LP formulation and re-solves using MPLP. We believe that our algorithm can use such an iterative approach as well at the expense of computational cost.

Table 1: Algorithm performance on 30 graphs of the Side-Chain Prediction data set.

|         | Optimal Solution | $\mu$(MAP error) | $\sigma$(MAP error) |
|---------|------------------|------------------|---------------------|
| DW-LP   | 16               | 7.51             | 24.18               |
| TRW-S   | 6                | 72.28            | 227.66              |
| MPLP    | 1                | 274.86           | 476.96              |
| MPLP-T  | 30               | 0                | 0                   |

Figure 1 displays the monotonic convergence property of our algorithm for graph 1ug6. At the last iteration, when the master program finds its optimal solution, the size of the master program consist of fewer than 70,000 variables. On the other hand, equation 5 for the same graph consists of approximately 650,000 variables; a significant reduction. The average size of equation 5 is approximately 460,000 variables (for the 30 graphs), the average size of the master program (upon convergence) is approximately 42,000 variables.

The implementation for TRW-S and MPLP methods are optimised C++ implementations from the respective authors' web-sites. Their average running times over all 30 graphs are: 2.70s, 37.16s, and 42.75s for TRW-S, MPLP, and MPLP-T respectively. Our algorithm is fully implemented in Matlab with the exception of the LP solver (CPLEX). The average running time for our algorithm is 46.49s when solving equation 5 directly (as all graphs fit in memory). When using the decomposition (equations 8 and 9) the average running time increases to 200.74s which is mostly due to additional overhead added in Matlab. The decomposition takes longer, but its strength is that it can handle large-scale problems with global constraints (which none of the other methods can) as shown in the next section.

## 5.2 Shape Matching

Shape matching can be seen as a data association problem where each point of a curve needs to be uniquely associated to another point of a different curve. In [26] Ramos et al. introduce a Conditional Random Field (CRF) for matching two sets of laser range finder data using only local shape information. Compared to the Side-Chain Prediction problem of section 5.1, this problem is not as difficult to optimise (the distribution is quite peaked). However, the scan matching problem is characterised by a very large state space if formulated as a LP relaxation, up to $362^{361}$ possible combinations. The reader is referred to [26] for further details on the CRF and its feature functions.

The quality of scan matching solutions improves if all nodes occupy a unique state (with the exception of a catch-all *outlier* state). In [26] LBP is used which does not allow for such constraints to be considered in the inference process. We therefore demonstrate the strength of our algorithm for dealing with both large-scale problems as well as external global constraints. Note that we are unable to apply TRW or MPLP methods as their implementations require all pair-wise potentials to be kept in memory. This exceeds the memory available on our test environment (4GB). Our LBP implementation overcomes the memory issue by recomputing the pair-wise potentials with each message sent.

In [26] 20 labelled data sets are used for training. Leave one out cross-validation is used on these 20 data sets to compare the quality of solution of LBP and the proposed algorithm. LBP will run for a maximum of 10 iterations or until convergence, whichever occurs first. The proposed algorithm will be run in a distributed environment. Subprograms are solved by a cluster of 5 machines with 4 or 8 CPU cores each. At each iteration all sub-optimal subprograms are added to the master program. Once the master program takes more than 2.5 seconds to solve, all non-basic columns are removed. Finally, the algorithm is run for a maximum of 250 iterations or until convergence. The comparisons are given in table 2 while figure 3 shows partial matching results for a single graph.

In figure 3 the blue crosses and red stars represent positions of an object measured by the laser range finder. The range finder will measure the same object but from different poses. The aim is to find unique matches between blue crosses and red stars represented by the black lines.

The results of table 2 show that our algorithm pro-duces a higher accuracy compared to LBP. The removal of many-to-one matches is particularly important for scan matching. However, LBP matches several points of an object in one set to the same point in the other set (figure 3), something not physically possible with rigid objects. Our DW-LP algorithm permits constraints on the solution, thereby eliminating all many-to-one matches and preserving object rigidity.

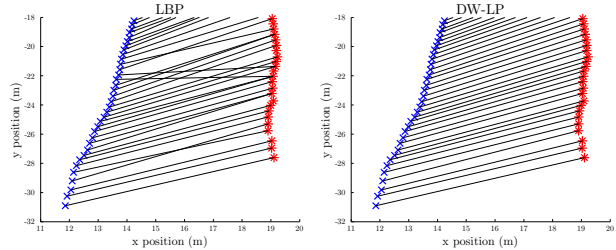

Figure 3: Scan matching results. Left hand figure show results from LBP, as can be seen several blue points (crosses) match to the same red point(s) (stars). The right hand figure is from the proposed algorithm, due to its global constraints it has no many-to-one matches. Note, black lines indicate a match between a blue and red point.

Table 2: Algorithm performance on 20 graphs of the scan matching data set.

|  | $\mu$(accuracy) | $\mu$(many-to-one matches) |
| --- | --- | --- |
| DW-LP | 44.99% | 0% |
| LBP | 28.88% | 16.16% |

## 6 Discussion

Many real-world problems are characterised, not only by the difficulty in solving them, but also by the size of the problem and constraints on its solution. Such problems require a different approach in algorithm design.

This paper presented a novel algorithm for distributed MAP inference based on LP decomposition. Unlike other LP (or QP) formulations ours is defined over edge variables instead of node variables. The advantage of such a formulation is that the LP has fewer constraints and allows decomposition into a number of subprograms (one for each edge) together with a small master program. The subprograms can be distributed over a network to allow large-scale problems to be solved efficiently. In addition, the master program monotonically converges to its optimal solution resulting in an anytime algorithm for performing MAP inference.

Experimental results show that the algorithm finds solutions comparable to current state-of-the-art and scales well to large problems. Additionally, the experiments showed that the algorithm can successfully be applied to problems that involve global constraints; a difficult task for message-passing based algorithms.